\documentclass[letterpaper, 10 pt, conference]{ieeeconf}
\IEEEoverridecommandlockouts   
\usepackage{cite}                       
\usepackage{amsmath,amssymb,amsfonts}
\usepackage{graphicx}
\usepackage{textcomp}
\usepackage{xcolor}
\usepackage{booktabs}
\usepackage{array}
\usepackage{tabularx}
\usepackage{float}
\usepackage{paralist}                   
\usepackage[hidelinks,breaklinks=true]{hyperref}
\hypersetup{pdfauthor={},pdftitle={},pdfsubject={},pdfkeywords={}}
\usepackage[capitalize]{cleveref}       
\usepackage{microtype}                  

\graphicspath{{figures/}}
\newcommand{\toolname}[1]{\begingroup\Urlmuskip=0mu plus 1mu\relax\nolinkurl{#1}\endgroup}

\crefname{figure}{Fig.}{Figs.}
\Crefname{figure}{Fig.}{Figs.}
\crefname{table}{Table}{Tables}
\Crefname{table}{Table}{Tables}
\crefname{section}{Section}{Sections}
\Crefname{section}{Section}{Sections}
\crefname{equation}{}{}
\Crefname{equation}{}{}

\makeatletter
\renewcommand\paragraph{\@startsection{paragraph}{4}{\z@}%
  {1.4ex \@plus .4ex \@minus .2ex}%
  {-0.55em}%
  {\normalfont\normalsize\bfseries}}
\def\@IEEEsectpunct{\ \,}
\makeatother

\newif\ifanonymous

\title{\LARGE \bf
DiagGen: Agentic Generation of Deformable Assets with Sim-based Diagnostics for Robotic Simulation
}

\ifanonymous
\author{Anonymous Author(s)}
\else
\author{%
Guanxiong Chen$^{1,3,*,\dagger}$, Yiduo Qu$^{2,*}$, Qianjun Xia$^{1,*}$,
Pengyu Jing$^{3}$, Yixian Cheng$^{1}$, \\
Bole Ma$^{4}$, Pengzhi Yang$^{3}$, Bingyang Zhou$^{3}$, Ziming Li$^{3}$, Shashwat Suri$^{1}$, Gongbo Sun$^{5}$,\\
Chao Liu$^{1}$, Peter Yichen Chen$^{1}$, Ziqiu Zeng$^{3,+}$, Fan Shi$^{3,+}$%
\thanks{$^{*}$Equal technical contribution.\quad $^{\dagger}$Project lead.\quad $^{+}$Corresponding authors.}%
\thanks{$^{1}$University of British Columbia, Vancouver, Canada.
        $^{2}$University of Cambridge, Cambridge, United Kingdom.
        $^{3}$National University of Singapore, Singapore.
        $^{4}$NHR@FAU, Friedrich-Alexander-Universit\"at Erlangen-N\"urnberg, Erlangen, Germany.
        $^{5}$University of Wisconsin--Madison, Madison, WI, USA.}%
}
\fi

\makeatletter
\newcommand{\titlegallery}{%
  \begin{minipage}{\textwidth}
    \centering
    \includegraphics[width=0.95\linewidth]{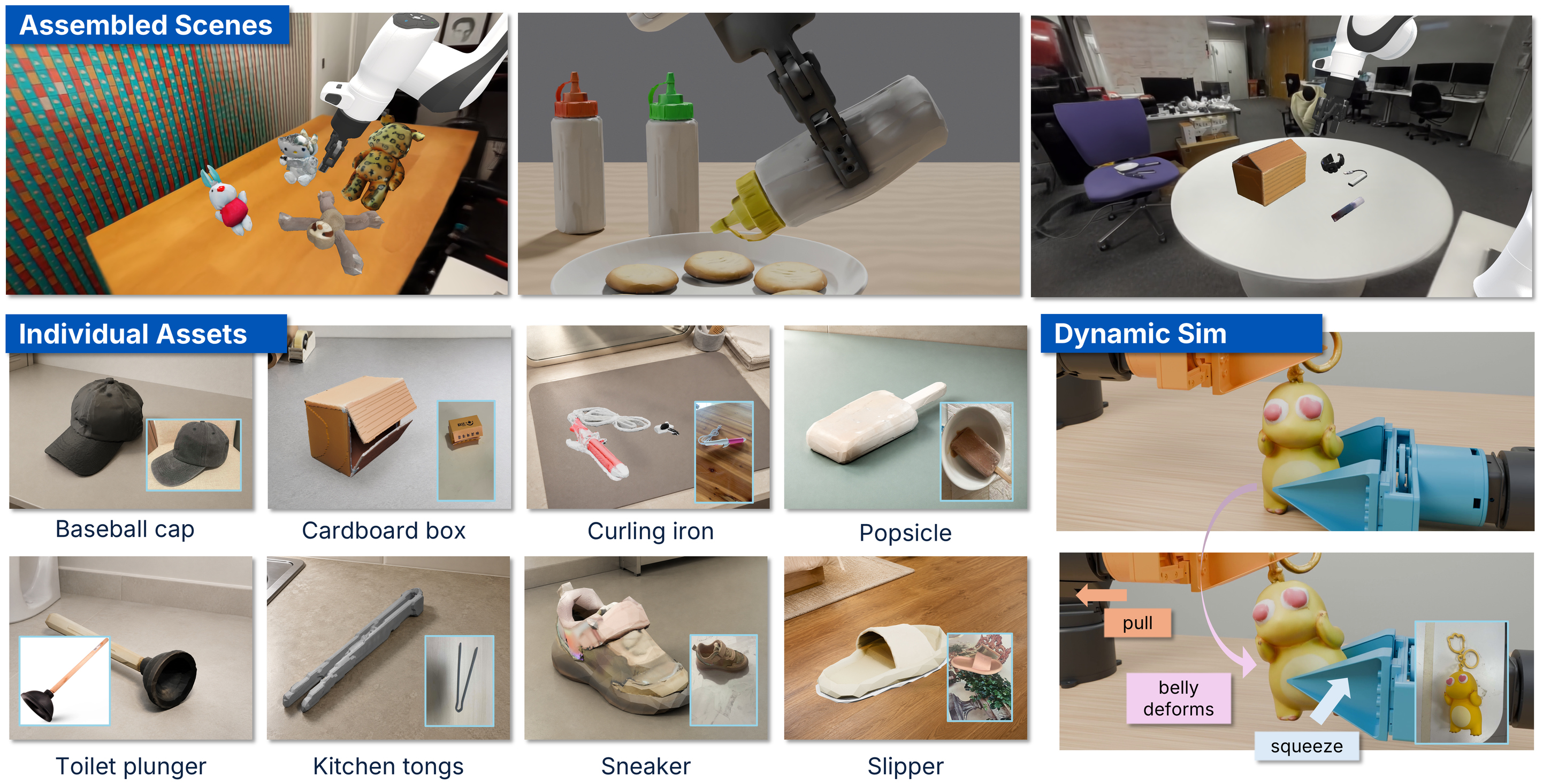}
    \par
    \def\@captype{figure}%
    \refstepcounter{figure}%
    \@makecaption{\fnum@figure}{\textbf{Selected DiagGen assets.} Top: scenes assembled from Gaussian-rendered backgrounds, DiagGen-generated deformables and assets from~\cite{calli2017yale}. Bottom: (Left) textured deformable assets, phone/internet image from which the asset was generated; (Right) dragon toy squeezed by gripper.}%
    \label{fig:asset_gallery}%
  \end{minipage}%
}
\makeatother
\IEEEaftertitletext{\titlegallery}

\begin{document}
\bstctlcite{IEEEexample:BSTcontrol}
\maketitle

\begin{abstract}
While simulation-ready deformable assets are essential for in-silico robotic manipulation tasks, existing generation frameworks typically assess physical plausibility after generation, leaving an object's simulated response unused as feedback for repairing upstream errors. We present DiagGen, an agentic framework that turns a single in-the-wild image into a simulation-ready deformable asset through a generate--simulate--diagnose--refine loop. DiagGen constructs part-aware geometry and material parameters, then uses a VLM (vision-language model)-based agent to select semantically informative regions, probe them in a physics simulator, observe material responses, and route evidence-backed repair cues to the responsible generation stage. Experiments on 40 assets show that diagnostics provides useful repair cues and can moderately improve the quality of generated deformable assets. Finally, we show that unlike assets generated from visual foundation models which may not be simulatable, DiagGen-generated deformables can be directly dropped into a high-fidelity physical simulator for the planning and simulation of contact-rich pick-and-place tasks. The project's website is \url{https://diaggen.github.io/}.
\end{abstract}


\begin{figure*}[t]
    \centering
    \includegraphics[width=0.98\textwidth]{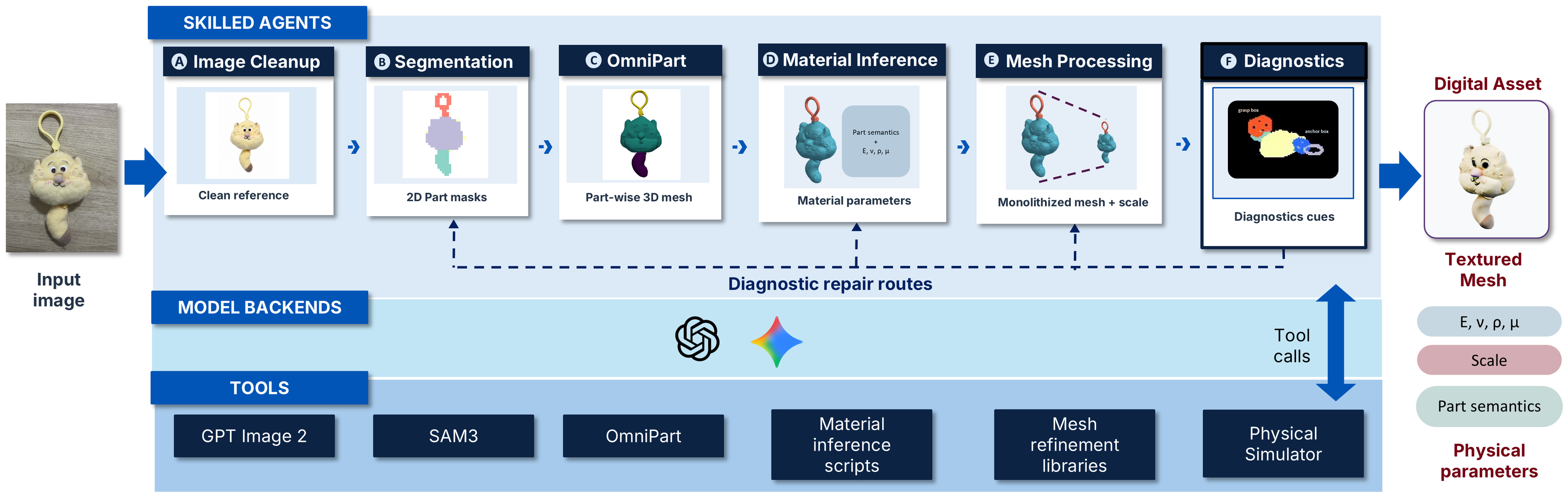}
    \caption{\textbf{Architecture of DiagGen.} DiagGen turns a phone photo into a simulation-ready deformable asset through a generate--simulate--diagnose--refine loop. The framework has three layers. On top, skilled agents at the top make high-level decisions for each stage and choose where a failed asset should be repaired. Directly below them, the model backends support the agents' reasoning. The tools at the bottom carry out routine operations, check intermediate results, and connect to the physics simulator. Diagnostics can send an asset back to an earlier stage for repair; accepted assets are exported for downstream applications.}
    \label{fig:architecture}
\end{figure*}

\section{Introduction}
\label{sec:intro}

Simulation-ready digital objects are essential for constructing diverse synthetic scenes and robotic manipulation data used to train embodied AI, and many recent works have sought to leverage visual-language models (VLMs) to generate such assets at scale~\cite{katara2024gen2sim, nasiriany2024robocasa, li2025urdf, zhou2026articraft, cao2026physx}. Frameworks such as PhysTwin, EMPM and DSO~\cite{jiang2025phystwin, chen2026empm, li2025dso, obrist2025pokeflex} focus on deformables, reconstructing simulatable soft objects from real-world robot-object interactions. These works leave two unaddressed limitations: First, agentic asset generation pipelines typically rely on ad-hoc validation methods to gauge the physical plausibility of assets; they do not make use of an individual deformable asset's in-simulator material response as a diagnostic signal to identify problems that lead to physically implausible behaviors, and revise the asset accordingly~\cite{zhou2026articraft,cao2026physx}. Second, deformable assets' material parameters are generally fitted through gradient-based optimization. Although effective for fitting a limited set of real-world observations, this formulation does not condition parameter inference on part semantics, and requires expensive calibrated interaction capture and per-object optimization.

To address these limitations, we make two design choices. First, building upon prior works which use VLMs or other foundation models to predict physical properties for individual parts~\cite{yang2025omnipart, yang2026physforge,cao2026physx}, we use VLM-guided, part-wise predictions to assign mechanical properties to different object parts. Second, we place a physics simulator inside the generation loop: unlike PhysX-Omni's post-generation tests, DiagGen tests each candidate through controlled interactions and uses the observed behavior to find the likely source of a problem~\cite{cao2026physx}, similar to how a human roboticist tests the usefulness of an asset. It then sends the asset back to the stage responsible for that problem. That stage rebuilds its output, and all later stages run again before the asset is tested once more.

We present \textbf{DiagGen}, an agentic framework that turns real-world images into simulation-ready deformable assets and uses tests in a physics simulator to find and repair problems. \Cref{fig:asset_gallery} shows several assets made from everyday photos; it also focuses on deformable objects, and beyond those covers heterogeneous objects with rigid parts, e.g. the curling iron. It asks two questions: Does a generated object behave realistically during a controlled interaction? Can information curated from a simulator guide a targeted repair? Our contributions are threefold:

\begin{compactenum}[C1]
    \item We introduce an agentic deformable asset generation workflow with simulation-based diagnostics in the loop. Given an in-the-wild image, stage agents generate geometry and material parameters, establishing a simulator-facing deformable asset. The diagnostic agent in the last stage commands a physics-based simulator to render visuals, applies grasps to probe regions of interest, and judges physical plausibility of the grasped regions from information obtained from the simulator.

    \item The diagnostic agent sends repair cues to earlier generation stages. In a 40-asset evaluation, these cues guide targeted changes to segmentation, material inference, and mesh processing, producing moderate improvements in the revised assets.

    \item We demonstrate various robotic utilities of DiagGen assets, showing that unlike unpolished assets generated by visual foundation models, DiagGen-generated deformables can be directly used for manipulation in a high-fidelity robotic simulator. Further, we show that more physically-realistic rollouts can be attained with an asset repaired following diagnostics, compared with its unrepaired counterpart; that a DiagGen asset can be situated in a synthetic scene transformed from the real world.
\end{compactenum}

\section{Related Work}
\label{sec:related_work}

\subsection{VLM-Driven Synthetic Data Generation}
Agentic, VLM-driven frameworks replace single-pass generation with a typical reason--act--observe--revise loop: LLM or VLM-driven agents plan content, invoke tools or programs, collect verifier evidence, and revise generated artifacts~\cite{pfaff2026scenesmith,yang2025sceneweaver,wang2026physcensis,kang2026simworldstudio,zhou2026articraft,ma2026lychsim,yu2025artgs,zook2025grs}. These frameworks differ mostly in whether they target object or scene-level reconstruction. Scene-level frameworks typically focus on revising layouts, views, lighting setups for better visual and physical realism of generated indoor scenes~\cite{pfaff2026scenesmith,yang2025sceneweaver,zook2025grs, ma2026lychsim, wang2026physcensis,kang2026simworldstudio, chen2026agentic}, whereas object-level frameworks often use CAD tools or foundation models to generate individual (articulated) rigid objects~\cite{zhou2026articraft,yu2025artgs,cao2026physx}. While DiagGen belongs to the latter agentic family with an object-level focus, its assets can be dropped into scene-level reconstruction frameworks such as Agentic Real2Sim~\cite{chen2026agentic} to build synthetic scenes that contain deformables.

\subsection{Generation and Simulation of Deformables}
Recent work~\cite{jiang2025phystwin, chen2026empm, zhou2026sim1, obrist2025pokeflex} focus specifically on the generation, simulation and manipulation of deformable assets and in particular, address several unique challenges posed by deformable assets: using mass-spring models for fast simulation and collision handling of high-DoF systems~\cite{jiang2025phystwin}, differentiable MPM for online modelling of elastoplasticity~\cite{chen2026empm} and leveraging high-fidelity scans for reconstructing accurate geometries~\cite{zhou2026sim1, obrist2025pokeflex}. We distinguish DiagGen from these earlier works in three respects: it starts from a single uncalibrated image, diagnoses and repairs a generated asset rather than reconstructing a specific real object from recorded real-world interactions~\cite{chen2026empm, jiang2025phystwin} or high-fidelity scans~\cite{zhou2026sim1}. Second, similar to ~\cite{obrist2025pokeflex} we use volumetric meshes as opposed to surface meshes~\cite{zhou2026sim1} to represent our assets, so assets can be accurately simulated in a high-fidelity FEM-based simulation such as Genesis~\cite{genesis2026genesisworld}. Third, we emphasize the heterogeneity of deformable assets: using VLM-supplied part-wise material parameters as in~\cite{cao2025sophy}, coupled with the linear elastic material model offered by Genesis, we can simulate deformables that contain relatively rigid parts in addition to soft parts.

\subsection{VLM-as-a-Judge for Open-Ended Evaluations}
Exact ground truths of real-world objects such as material labels are often unavailable, while collecting human judgments for every sample is expensive. These constraints have motivated the LLM-as-a-judge line of work, in which a model scores or compares outputs under a natural-language rubric~\cite{zheng2023judging, li2025generation}. To the best of our knowledge, no publicly available dataset provides part-wise material labels checked against real-world measurements of deformable objects; therefore a VLM judge is particularly useful, as it offers a practical way to compare simulated behaviors in the absence of human annotation. Earlier VLM-based data generation frameworks like Scenethesis and PhysX-Omni also use VLM judges to follow explicit grading guidelines when assessing rendered scenes and simulated physical behavior~\cite{ling2026scenethesis,cao2026physx}. Following this paradigm, we use two complementary evaluations: \textbf{Paired Asset Interaction Plausibility (PAIP)} compares the simulated behavior of baseline and revised candidates, while \textbf{Part--Material Specification Consistency (PMSC)} scores each asset's static part and material specification with a bounded correction for mesh cost. We detail both metrics in \cref{sec:metrics}.

\section{Method}
\label{sec:method}

\subsection{Overall Pipeline}
\label{sec:pipeline}

\Cref{fig:architecture} shows the architecture of DiagGen: it turns a single real-world image into a simulation-ready deformable asset through a generate--simulate--diagnose--refine loop. The generation path contains six stages. \emph{Image cleanup} converts the input image into a clean object reference and a simulation-oriented description that emphasizes rest shape and material cues. \emph{Segmentation} uses SAM3~\cite{carion2026sam} to decompose the reference into 2D parts, and \emph{OmniPart} lifts the masks and description into coherent part-separated meshes~\cite{yang2025omnipart}. \emph{Material inference} then assigns part-wise mechanical properties using a volumetric physical representation. \emph{Mesh processing} establishes metric scale, then combines the part meshes together into a single monolithic mesh, creating a loadable Genesis asset. The resulting candidate enters \emph{simulator-based diagnostics}; an accepted revision proceeds to final export together with its manifest, material specification, and diagnostic record.

Each stage consumes versioned upstream artifacts and emits a validated output pack, so a diagnostic recommendation can address the stage that owns the attributed fault. Agents use skills to describe the object, assign material parameters, define diagnostic anchors and grasp regions, make observations and reflect upon them, and assign repair routes. Low-level tools enforce artifact contracts and transform those choices into segmentation masks, meshes, simulator configurations, and probing actions. This boundary makes the workflow reproducible while preserving the open-ended reasoning needed for complex deformable objects.

\subsection{Simulator-Based Diagnostics}
\label{sec:diagnostics}

\begin{figure}[t]
    \centering
    \includegraphics[width=0.95\linewidth]{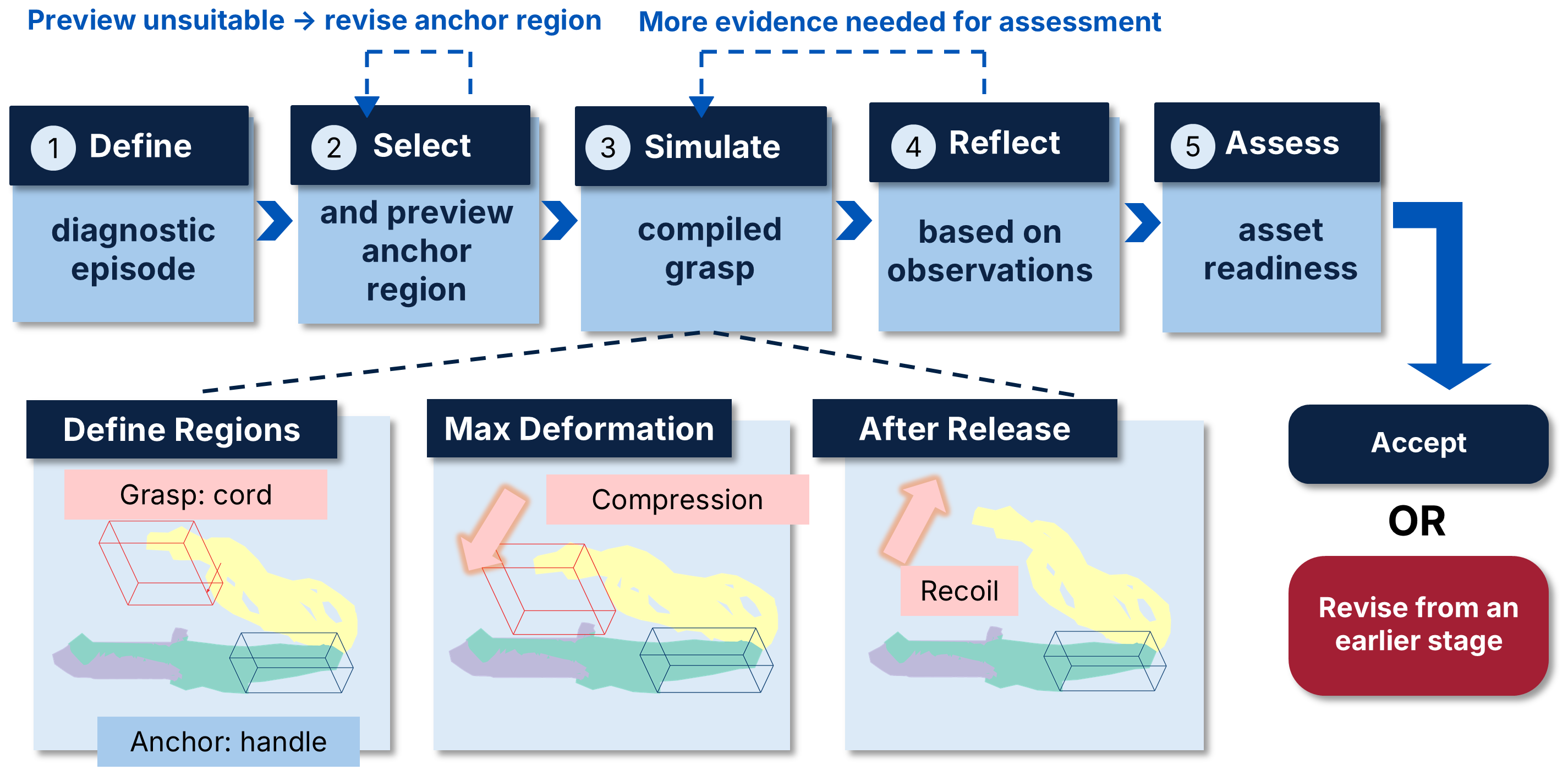}
    \caption{\textbf{Diagnostic-agent workflow.} If one grasp does not provide enough evidence, it runs another test in the same episode. The lower row shows an anchor and a grasp box, largest deformation upon compression, and a state after release observed by the diagnostic agent for further assessment of physical plausibility.}
    \label{fig:diagnostic_agent_flow}
\end{figure}

\Cref{fig:diagnostic_agent_flow} shows how the diagnostic agent tests a candidate and decides whether it is ready. The agent chooses what to test, creates a short diagnostic episode, previews the test region, runs a grasp in simulation, reviews the result, and decides whether more evidence or a repair is needed.

Diagnostics asks whether the candidate responds plausibly to interactions chosen for its geometry and semantics. The agent uses a scoped set of indexed parts, geometry summaries, and material specification to plan and ground regions of up to four semantic anchors, though the records do not by themselves determine the diagnosis. An anchor identifies a behaviorally informative region, such as a handle, rim, flap, corner, or compliant body; and it states the physical hypothesis to test. As shown in the lower row of \cref{fig:diagnostic_agent_flow}, the compiled grasp box pairs with the anchor, and specifies where an end effector should make contact, while the maximum-deformation and after-release panels show the motion used to expose and assess the relevant behavior. The runtime compiles grasp and anchor boxes and renders a preview. The agent then revises the target from the preview, until the contact location and motion direction express the intended test. We refer readers to the accompanying source code for the exact implementation of the diagnostic skills and tools.

The agent then begins a fresh simulation episode for each approved anchor. The episode begins from a reset state in which the object is at rest; an end effector then grabs the identified grasp region, and performs the prescribed displacement, and finally releases the object and allows it to settle. This protocol makes evidence comparable across revisions, while the selected anchor, displacement, duration, and observation schedule remain asset-specific. During probing, the agent receives synchronized RGB triptychs from top and oblique views, part-grounding overlays, geometry and action metadata, step summaries, and runtime logs. It can pause, observe, and issue additional probes when the current evidence leaves the physical hypothesis unresolved.

After each episode, the agent treats the observed response and simulator-supplied metadata as the primary diagnostic evidence. It then reflects upon the simulator-supplied geometry, material and simulation data to attribute any identified failure to geometry, scale, segmentation, representation, material response, discretization, or episode setup. Finally the agent aggregates its findings into a recommendation with supporting observations and a repair route. The orchestrator agent adjudicates the recommendation against the complete revision history and either accepts the asset, launches the selected repair, or halts the diagnostic process. Thus, the diagnostics agent acts as an online critic whose judgments are tied to controlled interactions and traceable evidence rather than a single rendered view.

\subsection{Evidence-Guided Repair}
\label{sec:repair}

DiagGen maps a diagnostic attribution to one of three repair routes: the \emph{segmentation route} revises part masks and boundaries when behaviorally distinct regions have been merged or a continuous region has been fragmented. The \emph{material-inference route} revises the surface-versus-volumetric representation and part-wise mechanical parameters when geometry is coherent but the observed deformation or settling is implausible. The \emph{mesh-processing route} revises scale, topology, or representation-specific geometry when contact and deformation expose a numerical or structural fault. Episode-setup findings remain inside diagnostics, where target and action previews are revised before additional evidence is collected.

A routed repair creates a new revision while reusing artifacts produced in earlier stages: segmentation preserves the cleaned reference; material inference preserves the reference, masks, and OmniPart meshes; and mesh processing additionally preserves the inferred representation and materials. The selected stage and every downstream stage are then re-executed, yielding a complete candidate rather than a local in-place edit. The new revision returns to the same diagnostic protocol, which enables direct comparison of anchor-level evidence across revisions. Once accepted, final export freezes the asset bundle and records the lineage from input image through repair decisions and simulator evidence.

\subsection{Evaluation Metrics}
\label{sec:metrics}

We define \textbf{Part--Material Specification Consistency (PMSC)} as a static assessment of the generated specification against the source image and common material priors. The score combines a whole-asset consistency judgment and a penalty for overly-complicated geometry. For each candidate, two independent VLM evaluators inspect the source image, segmentation, indexed part geometry, rest pose, part semantics, material class, Young's modulus $E$, Poisson's ratio $\nu$, density $\rho$, friction, fill mode, and metric mesh scale. Each evaluator sees one anonymous candidate, with diagnostic records, revision status, paired candidates, and the other evaluator's judgment hidden. The first evaluator assigns a whole-asset score $H$ from 1 to 5: 5 indicates strong support across the important static dimensions; 4 indicates an overall coherent specification with only minor local issues; 3 indicates mixed evidence with a supported concern; 2 indicates a major contradiction in an important functional region; and 1 indicates a fundamental contradiction in the object's representation or physical specification. The second evaluator identifies concerns by root cause and sets $F=1$ if any major concern undermines the representation or specification as a whole, and $F=0$ otherwise. The final score is
\begin{equation}
    S=\max\left(1,\,H-F-\mathbf{1}[T>40{,}000]\right),
    \label{eq:pmsc}
\end{equation}
where $T$ is the tetrahedron count, checked against the final mesh and its topology and scale summaries. The $40,000$-tetrahedron threshold is a heuristically-set computational budget to penalize overly complex meshes. Each correction subtracts at most one point.

Further, following the paired-comparison and order-swapping approach in~\cite{zheng2023judging}, and the visual-pair judging design of Scenethesis~\cite{ling2026scenethesis}, we define \textbf{Paired Asset Interaction Plausibility (PAIP)} as a \textit{dynamic} (simulation-based) testing regime to assess whether a diagnostics-guided revision improves the physical plausibility of an asset's behavior in the simulator over its unrevisioned baseline. As in PhysX-Omni~\cite{cao2026physx}, PAIP uses VLM to assess the plausibility of an asset's behavior in the simulator using visual evidence rendered by the simulator, rather than internal physical fields and asks for a judgment against explicit, human-interpretable plausibility criteria. For each revised asset, PAIP freezes one deformation probe---its target part, anchor, and action---from the baseline's diagnostic evidence before the revised candidate is seen, then replays it on both candidates under identical simulator, controller, and camera settings, with each candidate grounding the probe on its own parts. To reduce variability in evaluation, we set up two VLM judges, each of which independently compares the same two candidates twice: once as A vs. B, and once as B vs. A, since position bias is a well-documented limitation: a judge may favor the candidate presented first. Zheng et al.~\cite{zheng2023judging} and Li et al.~\cite{li2025generation} proposed swapping the presentation order as a mitigation. We use GPT-5.6 Terra at Max reasoning for the judges, different from the GPT-5.6 Luna at Max reasoning for generation and diagnostics in the diagnostics-and-repair experiment in Sec.~\ref{sec:diagnostic_guided_repair} to mitigate bias toward Luna-generated artifacts. Each judge then checks whether a target part responds as intended to a given grasp, whether the grasped anchor stays properly connected to the rest of the object, and whether the whole object behaves realistically. We then combine the four judgments: two judges, each seeing both candidate orders to decide whether the revised asset wins, the baseline wins, or the result is a tie.

\section{Experiments}
\label{sec:experiments}

\providecommand{\placeholder}[1]{\textcolor{red}{\textit{[#1]}}}

We evaluate whether diagnostics provides useful feedback for generating deformable assets and whether the resulting assets support downstream manipulation and scene-level digital twins. To examine C1 and C2, we ask:

\noindent\textbf{RQ1.} Can a VLM-driven agent autonomously author and execute asset-specific diagnostic episodes, and ground physical-plausibility judgments in evidence elicited through active simulator probing?

\noindent\textbf{RQ2.} Does routing simulation-grounded diagnostic cues to upstream stages improve the physical quality of regenerated assets?

\Cref{fig:asset_gallery} shows a subset of assets used for evaluation. We run the full pipeline on a fixed set of 40 objects, consisting of phone images sampled from the real world and prior work~\cite{han2024mvimgnet2}. For RQ1, we use three representative assets whose expected parts and material behavior allow us to inject faults. For RQ2, we compare two versions, before vs. after repair of each asset recommended by the diagnostics agent to go through revision. Both versions use the same raw image, upstream files, VLM settings, simulator, and evaluation rules. To complete the experiments in a reasonable time, we allow each run with diagnostics can make at most one repair.

\subsection{Seeded-Fault Diagnosis}
\label{sec:seeded_fault_diagnosis}

We first qualitatively evaluate whether the diagnostics agent can identify faults owned by different upstream stages from observed physical responses. For each of the three selected assets, we expose distinct physical errors, and map them one-to-one to the three repair route destinations. A single stage-owned issue per asset keeps the responsible stage unambiguous, while rerunning all subsequent stages allows the defect to propagate through the same artifact contracts used by the full pipeline. For the curling iron, we replace the flexible power cord's material parameters with the high stiffness assigned to the rigid barrel, producing a near-rigid cord under grasp-and-move interaction. For the massager, the generated segmentation captures only a small portion of the metal coil spring, preventing downstream stages from representing it as a complete, independently targetable part. For the kitchen tongs, we change the mesh-processing metric-scale target from 0.28~m to 2.8~m, producing an implausibly oversized tong mesh while leaving topology and material labels otherwise valid. These cases isolate material inference, segmentation, and mesh processing, respectively.

Each case resumes at the affected stage and executes every downstream stage; all diagnostics in this subsection use GPT-5.5 with high reasoning effort. To prevent leakage of the fault into context, the diagnostic agent receives no information about the issue or its originating stage. We freeze the remaining evaluation configuration, including probe and frame budgets, output schema, and retry policy, across cases. Before execution, a ground-truth record identifies the affected part, the issue, and its owning stage. Every diagnostic episode begins from a fresh reset, records a passive baseline, and applies the common approach--contact--displace--release protocol from \cref{sec:diagnostics}. Diagnostics evidence contains RGB views, target previews, part-grounding overlays, geometry and action metadata, step summaries, and runtime logs used by the agent. A case is correct when the agent both identifies the issue and routes the candidate to that stage. This joint criterion requires the elicited response to support an actionable diagnosis with a named issue and owning stage.

\begin{table*}[t]
    \centering
    \caption{\textbf{Seeded-fault diagnosis.} Each row presents the simulator observation first, followed by corroborating records and the resulting diagnosis. The final columns give the ground-truth issue and whether the trace identifies both the issue and intended route.}
    \label{tab:seeded_fault_diagnosis}
    \scriptsize
    \setlength{\tabcolsep}{3pt}
    \renewcommand{\arraystretch}{1.12}
    \begin{tabularx}{\linewidth}{@{}>{\raggedright\arraybackslash}p{0.12\linewidth}X>{\raggedright\arraybackslash}p{0.22\linewidth}>{\raggedright\arraybackslash}p{0.13\linewidth}@{}}
        \toprule
        \textbf{Asset} & \textbf{Simulation observation, corroboration, and diagnosis} & \textbf{Ground-truth stage and issue} & \textbf{Correct issue and route?} \\
        \midrule
        Curling iron &
        During probe, release, and settling, the cord/plug moved as a ``compact stiff assembly with little visible local bending''. The agent judged that ``resistance to local bending conflicts with the semantic prior for a power-cord component''. With targeting confirmed, it recommended \toolname{material_inference} to restore plausible cord flexibility during subsequent probe and release. &
        Material inference: barrel-like stiffness assigned to the flexible cord. &
        Yes; issue and route \\
        Massager &
        The agent judged that the probe lay on the coil, separate from the head-side anchor. It expanded and shifted the target along the ``visible coil span'', yet three attempts still failed to isolate the spring. It diagnosed a ``realized-region/foreground-resolution problem'' and requested \toolname{segmentation} to recover ``every visible helical turn and both coil-to-body boundaries'' as one continuous foreground spring region. &
        Segmentation: incomplete part mask omitted most of the metal coil spring. &
        Yes; issue and route \\
        Kitchen tongs &
        During the distal-tip probe, the mesh stayed connected while the ``distal arm bent substantially''. After release, two additional default-length unforced settling windows showed ``essentially no visible return toward the initial straight-arm configuration''. The agent judged the ``persistent post-release bend'' unsuitable for acceptance. The corroborating scale record identified an oversized mesh, so it recommended \toolname{mesh_processing} before reassessing material response. &
        Mesh processing: metric-scale target set to 2.8~m instead of 0.28~m. &
        Yes; issue and route \\
        \bottomrule
    \end{tabularx}
\end{table*}

\Cref{tab:seeded_fault_diagnosis} shows observations made by the diagnostics agent and its judgements: each simulator episode first exposes an implausible response or an inability to probe the intended component, and artifact records are then used to verify the target and discriminate among possible causes. For the curling iron, the material route connects near-rigid cable motion to improper stiffness assignment. For the massager, the segmentation route connects repeated failures to isolate the visible coil to its sparse realized ownership. For the kitchen tongs, the mesh-processing route connects persistent post-release bending to the corrupted metric-scale record. We can tell that across the three diagnostic cases, the agent uses asset-specific interaction evidence to correctly identify the affected component and physical discrepancy and to select the intended upstream stage, answering RQ1.

\subsection{Diagnostics-Guided Repair}
\label{sec:diagnostic_guided_repair}

For PMSC, we scored each of the 14 revised assets and its baseline separately using \cref{eq:pmsc}. We applied the rubric retrospectively to existing independent reviews. Mean PMSC rose from 2.14 before revision to 3.43 after revision, a moderate gain of 1.29 points on the five-point scale. Ten revised assets scored higher (71.4\%), four tied (28.6\%), and none scored lower. The remaining 26 were not revised as the diagnostics agent recommended acceptance on their first tries, and they scored an average of 3.12. Note that the diagnostics stage is not meant to repair most or even all assets generated, because that would imply pre-diagnostics agents fail to perform their jobs at a satisfactory level.

Turning our attention to the 14 assets for which the diagnostics agent recommends revision, we used PAIP to compare all 14 revised candidates with their baselines: the revised candidates win in nine cases, the baselines win in four cases, and for one pair we have a tie. The 9/14 outcomes reflect improvements in both access to the intended parts and their response to interaction. For the cookie, okra, and sneaker, the revised assets allowed the intended edge feature, distal tip, and narrow trim to be targeted and probed successfully, whereas their baselines failed to support those interactions. The hand brush showed a smaller gain: the probe selected the intended grip more precisely, while both versions remained connected and stable. For the toilet plunger, the revised rubber cup responded more strongly relative to the handle under the same force, better matching its expected flexibility, although the baseline recovered more cleanly after release. But repair did not improve every comparison: The baseline plush kitten allowed slightly better selection of the garment and produced a similar response with less force. The baseline sport watch showed a clearer response of the buckle and keeper while remaining connected to the housing; the revised target barely moved. These cases show why a completed repair still needs interaction testing: a change can improve one aspect of the asset without improving the intended physical response.

We next examine how diagnostic cues change upstream generation decisions. In \Cref{tab:diagnostic_repair_traces}, we report three independent cases, one for each of the three repair destinations considered in our pipeline. Each row follows the same evidence hierarchy: an observed simulator response or initialization failure that exposes the defect, artifact records that leads the diagnostics agent to root cause, diagnostics cues provided to the owning stage, and how generation strategy is changed by the owning stage before rebuilding the candidate. 

\begin{table*}[t]
    \centering
    \caption{\textbf{Diagnostic cues and upstream strategy changes.} Each row begins with evidence exposed by a simulator episode and then reports the corroborating records, routed stage, and corresponding change in upstream generation strategy.}
    \label{tab:diagnostic_repair_traces}
    \scriptsize
    \setlength{\tabcolsep}{3pt}
    \renewcommand{\arraystretch}{1.12}
    \begin{tabularx}{\linewidth}{@{}>{\raggedright\arraybackslash}p{0.10\linewidth}X>{\raggedright\arraybackslash}p{0.16\linewidth}X@{}}
        \toprule
        \textbf{Asset} & \textbf{Diagnostic cue excerpt} & \textbf{Routed stage} & \textbf{Repair-agent strategy-change excerpt} \\
        \midrule
        USB hub &
        From the static triptych, the agent judged the terminal outside the pinned cable-root support. Expanding the target still failed to isolate the plug. It diagnosed a ``missing visual-ownership/foreground-assignment defect'' and requested segmentation to restore the silver terminal as a distinct visible region. &
        Segmentation &
        The agent replaced generic USB-part descriptions with explicit regions for the housing, cable, connector, and metal plug. After regeneration, the plug appeared as a distinct component that the simulator could target reliably. \\
        Juice box &
        At the start of the diagnostic episode, Genesis rejected the asset before reset, so no probe could run. The runtime log reported local self-clearance below the assigned shell-thickness threshold; the material record then confirmed the incompatible uniform shell specification, and diagnostics routed the failure to material inference. &
        Material inference &
        The agent reduced the per-part shell thicknesses and replaced a single 3.0~GPa stiffness with distinct values of 0.8, 0.7, 0.6, and 0.3~GPa for the body, top, seam, and flap. The rebuilt asset initialized, completed probe and release tests, and was accepted. \\
        Medical bag &
        During probing and release, the agent saw ``visually stable triptychs'' of the surviving pouch body and front patch. Yet the expected straps, handles, and hardware were unavailable for interaction, leaving an incomplete simulated bag despite its stable body motion. Grounding confirmed the missing regions, prompting repair through mesh processing. &
        Mesh processing &
        The agent increased the meshing fidelity of the missing parts. The rebuilt connected mesh restored all five part labels with 13,318 tetrahedra and passed topology validation. \\
        \bottomrule
    \end{tabularx}
\end{table*}

These results suggest that diagnostics can identify artifacts in an open-loop generated asset, and provide useful guidance for improving the overall quality of a generated asset.

\subsection{Robotic Application: Case Studies}
\label{sec:robotic_app_pick_n_place}

\begin{figure}[t]
    \centering
    \includegraphics[width=0.95\linewidth]{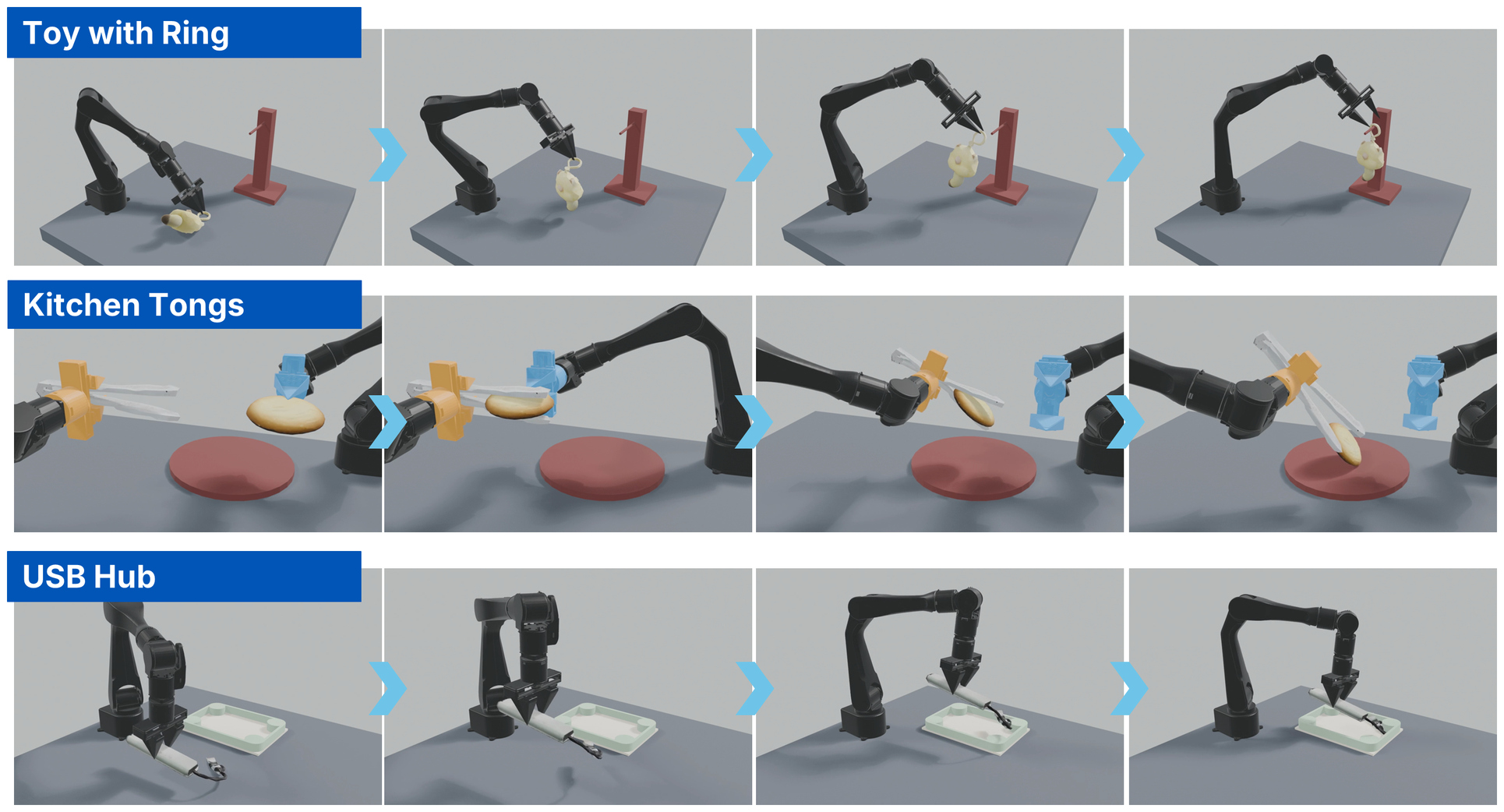}
    \caption{\textbf{Trajectory optimization for pick-and-place.} We conducted trajectory optimization for pick-and-place of three generated assets: a soft toy with rigid keyring on top, kitchen tongs and cookie, and a USB hub. Each row shows the optimized pick-up, lift, transport, and place sequence from left to right.}
    \label{fig:robotic_pick_and_place}
\end{figure}

Following PhysTwin's use of reconstructed deformable twins as dynamics models for model-based robot planning~\cite{jiang2025phystwin}, we evaluate whether a DiagGen asset can similarly support a downstream manipulation task. We use the implicit FEM solver provided in the Genesis simulator~\cite{genesis2026genesisworld}, with SAP for accurate frictional contact between the gripper and the deformable asset~\cite{han2023convex}. We perform task-space trajectory optimization with a GPT-5.6 Sol agent at extra-high reasoning. For each asset, the agent iteratively optimizes the gripper pose and aperture: it proposes a candidate sequence of actions, simulates and inspects visual observations, and reasons about a better grasp before resimulating, repeating this loop until the gripper achieves a firm grasp and holds the asset. Then it optimizes the transport and placement trajectory in the same fashion. \Cref{fig:robotic_pick_and_place} shows the resulting pick-and-place sequences for three generated assets. The toy with keyring and kitchen tongs are generated successfully on the first attempt, while the USB hub is revised from the segmentation stage.

\begin{figure}[t]
    \centering
    \includegraphics[width=0.95\linewidth]{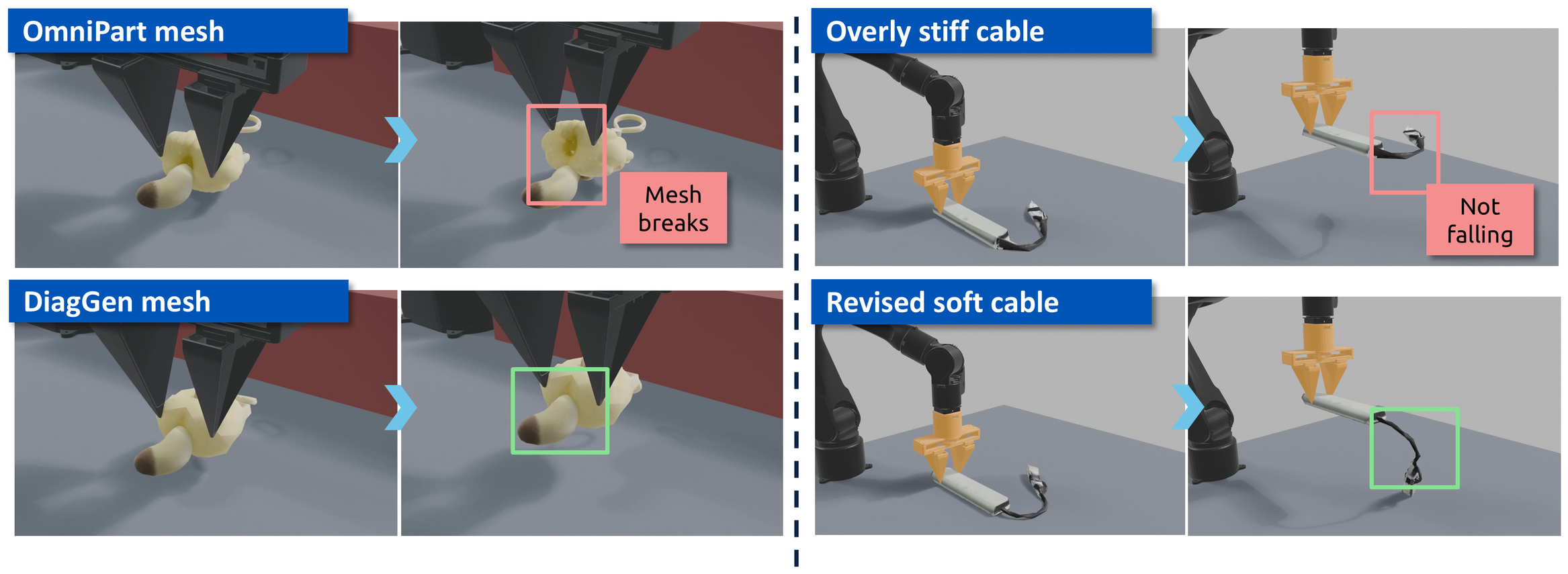}
    \caption{\textbf{Improved structure and material heterogeneity matter for manipulation.} Left: Under the same grasp-and-lift action, the tetrahedralized OmniPart toy mesh separates because its disconnected parts lack an elastic load path, whereas the monolithized DiagGen mesh stays coherent. Right: One Airbot gripper grasps the USB hub case and lifts it from the support surface. The baseline and revised assets begin from similar poses. During the lift, the baseline cable retains an overly stiff bend, while the revised cable deforms naturally and lets the connector hang downward. Colored boxes mark the affected regions.}
    \label{fig:probegen_comparisons}
\end{figure}

\textbf{Comparison with assets generated directly by OmniPart.} We ran the same grasp-and-lift action in Genesis on two toy-with-ring (Cheetah) meshes: a tetrahedralized OmniPart reconstruction and a mesh processed by DiagGen. As shown on the left of \Cref{fig:probegen_comparisons}, the OmniPart mesh breaks into separate parts because it is not monolithized and therefore cannot behave as one deformable object; the DiagGen mesh on the other hand stays coherent because monolithization connects its semantic regions into one deformable assembly,
and it can therefore be grasped and lifted as a single object. This comparison shows how DiagGen turns a disconnected reconstruction into an asset that supports robotic manipulation in a high-fidelity simulator.

\begin{figure*}[t]
    \centering
    \includegraphics[width=0.95\textwidth]{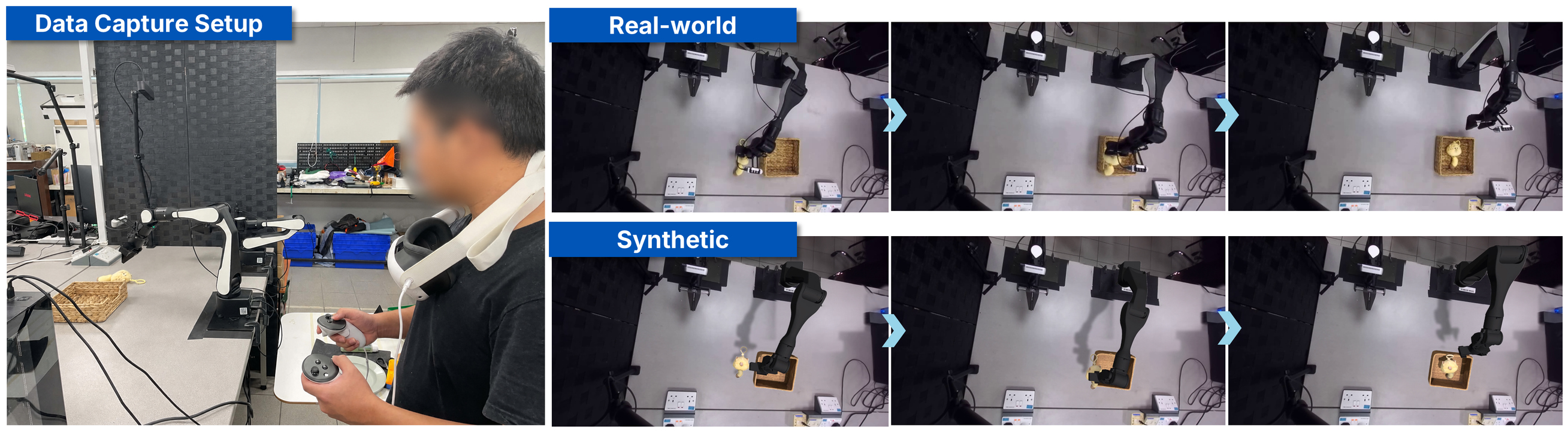}
    \caption{Using the Cheetah asset for scene-level reconstruction and trajectory replay. (Left) Experiment setup; (Right, top) real-world trajectory and (Right, bottom) simulated twin.}
    \label{fig:robotic_scene_real2sim}
\end{figure*}

\textbf{Effect of repair targeting segmentation.} For the USB hub in \cref{tab:diagnostic_repair_traces}, we compare the initial version with the version produced after diagnostics-guided segmentation repair under the same one-arm grasp-and-lift action. An Airbot gripper grasps the case and lifts the hub from the support surface. In the initial version, poor segmentation prevents the cable from being represented as a distinct deformable part, so downstream processing assigns one material to the whole asset. The cable therefore retains an overly stiff bend as the case rises; yet after repair, DiagGen can assign different material behavior to the cable and case. Under the same action, the cable stays connected, deforms under gravity, and lets the connector hang naturally, as shown on the right of \Cref{fig:probegen_comparisons}. The repaired asset therefore produces a more plausible response during manipulation.

\textbf{Scene-level Real2Sim.} We demonstrate another use case of a DiagGen-generated asset in which it is being used by a scene-level real2sim framework for converting a real-world deformable manipulation episode into its digital twin. Fig.~\ref{fig:robotic_scene_real2sim} showcases the experiment setup using an Airbot arm and gripper~\cite{airbot2026} and Meta Quest 2~\cite{metaquest2} for teleoperation; the real-world pick-and-place trajectory, as well as its synthetic twin generated by Agentic Real2Sim~\cite{chen2026agentic}. With the SAM3D-generated~\cite{chen2026sam} asset replaced by the DiagGen-generated asset and the MuJoCo backend replaced by Genesis, which is better suited to deformable simulation, Agentic Real2Sim is able to faithfully reconstruct the episode.

\section{Conclusion, Limitation and Future Work}
\label{sec:conclusion}

We presented DiagGen, a framework for generating simulation-ready deformable assets from a single in-the-wild image. It combines part-aware geometry and material construction with active simulator probing, allowing a diagnostic agent to make use of a high-fidelity physical simulator to identify implausible physical responses, inspect dimensional correctness, reason about which earlier stage is problematic, and send repair cues to the responsible generation stage. The seeded-fault experiments qualitatively show through three representative examples that the agent can identify errors in segmentation, material inference, and mesh processing and attribute them to the correct owning stage. In the 40-asset evaluation, diagnostics led to 14 revisions, of which 10 received higher PMSC scores under retrospective scoring. These results support diagnostics as a useful source of feedback that can moderately improve generated assets. The pick-and-place and scene-level real2sim demonstrations further showed that the resulting assets can support contact-rich robotic simulation.

DiagGen has two main limitations. First, failures in material inference and segmentation can produce similar symptoms, so the diagnostic agent may struggle to identify the root cause. Second, large-scale, parallelized data generation has not yet been tested. Future work will expand asset and interaction-data generation, measure how asset diversity and physical fidelity affect policy training and manipulation success, and extend DiagGen to articulated assets.

\section*{Acknowledgment}

We thank Beichen Li from OpenAI for support with computational resources.


\bibliographystyle{IEEEtran}
\bibliography{main}

\end{document}